\documentclass{article}

\usepackage{arxiv}
\usepackage[utf8]{inputenc} 
\usepackage[T1]{fontenc}  
\usepackage{hyperref} 
\usepackage{url}   
\usepackage{booktabs} 
\usepackage{amsfonts} 
\usepackage{nicefrac} 
\usepackage{microtype}   
\usepackage{graphicx}

\usepackage[sort,numbers]{natbib}

\title{Line Goes Up? Inherent Limitations of Benchmarks for Evaluating Large Language Models}

\author{
 James Fodor \\
 The Centre for Brain, Mind and Markets \\
 Faculty of Business and Economics\\ 
 The University of Melbourne, Australia \\
 \texttt{Fods12@gmail.com} \\
}

\begin{document}

\maketitle
\begin{abstract}
Large language models (LLMs) regularly demonstrate new and impressive performance on a wide range of language, knowledge, and reasoning benchmarks. Such rapid progress has led many commentators to argue that LLM general cognitive capabilities have likewise rapidly improved, with the implication that such models are becoming progressively more capable on various real-world tasks. Here I summarise theoretical and empirical considerations to challenge this narrative. I argue that inherent limitations with the benchmarking paradigm, along with specific limitations of existing benchmarks, render benchmark performance highly unsuitable as a metric for generalisable competence over cognitive tasks. I also contend that alternative methods for assessing LLM capabilities, including adversarial stimuli and interpretability techniques, have shown that LLMs do not have robust competence in many language and reasoning tasks, and often fail to learn representations which facilitate generalisable inferences. I conclude that benchmark performance should not be used as a reliable indicator of general LLM cognitive capabilities.
\end{abstract}


\section{Introduction}
The recent development of large language models (LLMs)\footnote{A note on terminology. In this article I use the term ‘large language models’ (LLMs) to refer to models based on the transformer architecture trained on large corpus of text data to interact with the user via natural language. Examples of such models include the Llama series, the GPT series, and the Gemini series. Although these models are capable of tasks beyond language processing, such as mathematical and coding tasks, this is still performed by next token prediction after training on linear strings of text. I avoid the term ‘artificial intelligence’, as it has no accepted clear meaning and does not refer to any specific model architecture.} based on the transformer architecture has led to extensive discussion as to how to best measure their capabilities. The most common way to assess LLMs is by their performance on standardised tests called benchmarks. It is often argued that recent substantial improvements in LLM benchmark scores, largely driven by state-of-the-art systems like GPT-4 and OpenAI’s o3, are indicative of large increases in the capability of LLMs. On this basis, it is often inferred that such models are becoming significantly more capable, potentially outstripping human-level performance, on a wide range of real-world tasks. In this article I contend that this argument is flawed in two key respects. First, I argue that inherent limitations with the benchmarking paradigm, along with specific limitations of existing benchmarks, combine to render benchmark performance highly unsuitable as a metric for generalisable competence across a range of tasks. Second, I argue that an alternative set of methods more suited for investigating the robustness and generalisability of LLM capabilities, namely adversarial stimuli and interpretability techniques, has yielded strong evidence that LLMs do not have robust competence in many language and reasoning tasks, and often fail to learn representations which facilitate generalisable inferences.

\section{Problems with benchmarks}

A benchmark is a test designed to assess the ability of an LLM to perform a particular type of task. They are designed to interrogate the ability of LLMs to perform tasks such as grammatical judgements, commonsense reasoning, logical inference, answering science or math questions, and producing computer code. Hundreds of benchmarks have been developed over the past decade, and it is outside my scope to review them here. Interested readers can consult existing literature reviews \citep{chang2024survey, ivanov2024ai, minaee2024large}. Here my purpose is to highlight several major limitations inherent in the practise of using benchmarks to assess the real-world capabilities and competence of LLMs, along with specific limitations of existing popular benchmarks. Critically, benchmarks aim to assess LLM capabilities in performing certain types of tasks that are similar, but not identical, to those contained in the benchmark itself. A high benchmark score is therefore not meaningful unless we have reason to believe the benchmark provides a good estimate for performance on related real-world tasks. In this section, however, I will argue that this is rarely the case. 

\subsection{Over-fitting to the benchmarks}

A major problem with many benchmarks is that shortly after they are made publicly available, they become incorporated into the data used to train the very LLMs they are intended to assess. Modern LLMs are easily large enough to memorise large swaths of data, with numerous studies finding evidence of benchmark tasks having been learned during model pre-training or subsequent fine-tuning \citep{laskar2024systematic, satvaty2024undesirable, wang2024generalization}. Further, even if LLMs do not directly memorise the answers to a task, exposing LLMs to example problems from a specific benchmark allows them to learn statistical regularities associated with that task. This can allow the LLM to achieve a higher score on that benchmark without any actual improvement in the LLM's general reasoning capabilities \citep{laskar2024systematic}. 

More broadly, the development of LLMs has become guided by benchmark performance to the extent that the benchmarks lose much of their value in assessing LLM capabilities. This problem is an instance of Goodhart's law, the adage that “when a measure becomes a target, it ceases to be a good measure”. Initially presented in the context of economics, it also applies in the context of evaluating LLMs \citep{banerjee2024vulnerability, thomas2022reliance}. Unfortunately, its importance in this context often goes unappreciated. For instance, recently the World Economic Forum erroneously inferred that LLMs have exceeded human performance on various cognitive tasks, on the basis that LLMs have exceeded human performance on benchmarks designed to test those task \citep{WEF2024}. Such an inference is only valid if the benchmarks in question are unbiased and representative indicators of LLM performance for the corresponding task as a whole. This measurement function of benchmarks, however, has been supplanted by their use as targets. New LLMs are designed specifically to achieve ‘state-of-the-art performance’ on various benchmarks, with this being rewarded by conference papers, citations, and media attention. Such incentives result in model architectures, parameters, and training regimes being fitted specifically to these benchmarks, thereby reducing their relevance for assessing LLM performance \citep{alzahrani2024benchmarks}. Such over-fitting to the test is manifested in the rapid pace with which new benchmarks become saturated, even while fundamental limitations of LLM competence are still evident, thereby requiring still newer benchmarks to assess performance \citep{reuel2024betterbench, banerjee2024vulnerability}.

Similar problems have long been appreciated in the field of psychometrics \citep{devon2007psychometric}. When people practise or receive coaching on components of an IQ test, performance on that task improves \citep{bartels2010practice}, but the test become less predictive of subsequent performance on other tasks \citep{te2001practice}. This phenomenon results in recommendations to limit the number of times a given test is assigned to a particular individual. The problem is likely to be even worse in the case of LLMs, as compared to humans they can be much more thoroughly adapted to specific tasks through architectural design choices, hyperparameter selection, choice of training data, and fine-tuning paradigm. This further highlights the difficulty of using benchmarks as both research targets and as evaluation metrics.

Supplementing these theoretical concerns, several recent studies have analysed this phenomenon empirically. One study found that fine-tuning various LLMs on the training portion of question answering and text comprehension datasets resulted in a doubling of performance. They also found that fine-tuning on some benchmarks led to a decline in performance on other benchmarks \citep{zhou2023don}. Another study found that leading LLMs such as GPT-4 had been exposed to many commonly-used public machine learning datasets in its training data. Indeed, in five out of seven cases GPT-4 had learned the data sufficiently well to be able to accurately sample from the dataset and reconstruct conditional distributions. These results highlight the difficulty of assessing LLMs using benchmarks which were publicly known at the time of their development \citep{bordt2024elephants}.

\subsection{Relevance of benchmark tasks}

Another significant problem with existing LLM benchmarks is the lack of research concerning their behavioural relevance. Few benchmarks provide substantive theoretical justification concerning how or why certain tasks were selected, or empirical evidence that strong performance on the benchmark correlates with performance on tasks of real-world relevance. While psychometrics has grappled with these questions for decades and developed sophisticated (though still imperfect) methods for addressing them, theoretical and empirical evaluation of LLM benchmarks lags far behind \citep{federiakin2025improving, reuel2024betterbench}. Too often, it is simply assumed that if a task requires intelligence for a human to perform then it will also be useful as a measure of the cognitive capabilities of an LLM. Decades of artificial intelligence research, however, has found that many tasks which require intelligence and flexible cognitive capabilities for humans to perform can nonetheless be accomplished by artificial systems that are not intelligent and lack general cognitive capabilities. Examples of this phenomenon, sometimes called Moravec's paradox \citep{hassabis2017artificial}, include executing search and sorting algorithms, automated theorem proving, playing games like chess and Go, competing in jeopardy, and many natural language tasks. Given the lack of careful research and poor track record of prediction, there is good reason to be skeptical about how informative current LLM benchmarks are about generalisable cognitive capabilities.

Survey evidence from users and developers of LLMs highlights this gap between benchmarks and real-world applications. Recently a series of interviews was conducted with 19 policy analysts, academic researchers, and industry professionals who have used benchmarks to inform decisions regarding adoption or development of LLMs \citep{hardy2024more}. Most respondents were skeptical of the relevance of benchmark performance for real-world tasks, as they found it difficult to relate questions on the benchmarks to tasks of real-world relevance or customer requirements. One especially blunt respondent reported: ‘Some of the major benchmarks that people use today, MMLU, for example, (if you) just go through the questions and look at them, and you’ll realize that they are awful… a bunch of totally irrelevant stuﬀ, things that are totally wrong.’

An additional factor limiting real-world applicability is that many benchmarks suffer from poor quality control and inadequate documentation of procedures for assessing accuracy of data. For example, one analysis found that 30\% of Google’s emotion dataset is mislabelled \citep{chen2022}. An analysis of NLI benchmarks similarly found that many of the questions either had incorrect answers or were too vague to have a single objective solution \citep{Kalouli2023NLI}. Similar problems have been documented for the MMLU benchmark \citep{gema2024we}. Furthermore, multiple prompts taken from a single benchmark commonly exhibit highly correlated performance across different LLMs, driven in part by semantic similarity between prompts. This indicates that benchmarks do not provide random sampling of problem types, which increases the risk that over-fitting to the benchmark will result in failure of generalisation \citep{siska2024examining}. Overall, quality control is clearly on ongoing challenge for LLM benchmarks.

\subsection{Analysis of recent benchmarks}

In an effort to rectify some of these problems, recently several new benchmarks have been developed which aim to provide a more rigorous test of LLM capabilities. Several of these benchmarks have recently attracted attention due to OpenAI’s new reasoning models o1 and o3 achieving substantial performance gains on these tasks. Here I consider several of these newer benchmarks in more detail, discussing the extent to which they successfully resolve the problems highlighted above.

SciEval is a benchmark consisting of 18,000 questions covering physics, chemistry, and biology, which utilises dynamically generated data to reduce the problem of contamination of training data \citep{sun2024scieval}. This dynamic component of the benchmark highlights just how extensive data contamination is, with GPT4’s accuracy on physics questions of falling from 65\% on publicly available static data to 26\% on dynamic data. While the use of dynamic data is a step forward, it is likely to be insufficient to resolve the problem of data contamination, since LLMs can learn not only the precise numerical values but also the form and structure of the questions \citep{li2024task}. This so-called task contamination has been found to be responsible for about a 20 percentage point increase in performance of the GPT-3 series models when comparing older to newer benchmarks.

The GPQA benchmark consists of multiple-choice questions covering chemistry, physics, and biology which are said to be ‘Google-proof’ \citep{rein2023gpqa}. This claim is based on the fact that performance of non-experts on the questions only marginally increased when they were given access to the internet. The questions are also claimed to be difficult, as domain experts achieved 65\% accuracy compared to a mere 34\% (on a 25\% baseline) achieved by non-experts with unrestricted internet access. However, the authors also found that LLMs achieved about the same score on the most difficult subset of questions as on the easier subset of questions, which contrasted sharply with the observation that human experts performed four times as well as non-experts on the difficult questions compared to only twice as well on the easier questions. This discrepancy indicates that LLMs are likely performing the task in a different manner to humans, raising the question as to whether they have truly obtained generalisable competence in the relevant subjects, or instead are relying on superficial heuristics to answer questions.

The FrontierMath benchmark is comprised of novel math problems developed specifically for the purpose by domain experts \citep{glazer2024frontiermath}. To prevent contamination of training data, it has not been made publicly available. While these are important steps forward, the authors still fall prey to the common mistake of assuming without evidence that their benchmark succeeds in testing the construct it was designed to test. Specifically, they claim that “the problems in FrontierMath demand deep theoretical understanding, creative insight, and specialized expertise”. While the novel math problems may well require deep theoretical insight for \textit{humans} to solve, it is an entirely open question whether they require equivalent understanding or insight for \textit{LLMs} to solve. This is especially relevant given that all tasks on the benchmark have numerical answers, meaning that correct reasoning is not assessed, only the solution. As such, it is unclear whether this benchmark will do much better than ensuring that models which achieve high scores have genuine mathematical competence, instead of merely utilising complex but superficial heuristics for guessing the answers.

RE-Bench consists of seven machine learning assignments where performance is evaluated using metrics such as the accuracy or efficiency of the resulting code \citep{wijk2024re}. Unlike most other benchmarks, RE-Bench requires the LLM to generate workable code, rather than simply output a multiple-choice option or numerical answer. While undoubtedly a useful resource, this benchmark does not live up to its claim of testing the capabilities of LLMs to serve as `AI researchers', as it only assesses their ability to perform modestly-sized, clearly-defined, non-interacting coding tasks. Another major limitation is that while producing their answers, LLMs and humans were both able to submit preliminary solutions and receive scores on a test set as live feedback. While the 8-hour time limit meant that it was difficult for humans to make much use of this information, the LLMs were able to quickly iterate many slightly different solutions to effectively brute-force increased scores against the test set. The authors also note that several times LLMs failed to adhere to the task instructions, which led them to produce solutions which performed well on the automated performance evaluation but were manually marked as incorrect for not adhering to the task. This raises the question of how often LLMs are able to achieve nominally correct answers despite following incorrect reasoning processes on other benchmarks where such manual inspection is not possible.

Overall, while these novel benchmarks represent a step forward for evaluation of LLMs, I conclude that they have failed to mitigate the major problems of data contamination, lack of real-world relevance, and poor evidence for generalisability that have plagued previous LLM benchmarks.

\section{Generalisation abilities of LLMs}

The purpose of benchmarks is to provide metrics of LLM performance which generalise beyond the specific questions contained in the benchmark to a range of tasks of real-world relevance. There are several aspects of successful generalisation, including performance of tasks from outside the training distribution \citep{zhang2023paradox}, ability to combine previously learned components in novel ways (compositionality) \citep{yao2022structural}, and robustness to task alternations such as noise, changes in format, and irrelevant information \citep{ullman2023large}. In the previous section I argued that current benchmarks are poorly suited for the purpose of measuring generalisable LLM performance on cognitive tasks. In this section I argue that we have strong evidence from adversarial, interpretability, and capability research that current LLMs are unable to learn to perform many cognitive tasks in a robust and consistent manner. 

\subsection{Adversarial stimuli and tasks}

One useful technique for interrogating the capabilities of LLMs is to develop tasks or stimuli which are deliberately designed to be challenging for them. Such adversarial tasks are analogous to tasks developed in the heuristics and biases literature to study the limitations of human cognition \citep{west2008heuristics}. In this section I highlight several recent studies which have bearing on the question of how well LLMs are able to learn to generalise beyond the problems they were trained on.

In one study focused on automated evaluation of LLMs, researchers developed a simple adversarial model which always produced the same output in response to all instructions \citep{zheng2024cheating}. However, this adversarial model was cleverly designed to reliably fool automated LLM evaluation software by manipulating the format of its output. This technique was able to fool even advanced evaluation models like GPT-4 about 80\% of the time. The inability to detect or appropriately respond to what humans would easily recognise as an obvious attempt to circumvent the test highlights the lack of any human-level understanding of the input. Such fragility also indicates that robust generalisation to novel styles and formatting of responses is likely to be difficult.

Other studies have used reversed text as an adversarial stimulus. This is motivated by the idea that if LLMs learn generalisable language competence analogous to that possessed by humans, they should show much greater competencies when trained on forward compared to reversed text. One prominent example relates to a widely publicised study which found that LLMs performed better than human experts at predicting the outcomes of neuroscience studies based only on their abstract \citep{luo2024large}. A follow-up study replicated this effect with models trained on character-reversed neuroscience articles, even showing slightly better performance than models trained on regular text. These results indicate that LLMs do not perform the task by extracting the same sorts of features that humans use in assessing the content of a scientific article \citep{luo2024beyond}. Sensitivity to stimulus structure has also been investigated at the syntax level, with one study finding that ChatGPT performs better than humans at judging grammatical sentences, but when asked about ungrammatical sentences only achieves about 35\% accuracy compared to humans 75\% \citep{dentella2023systematic}. ChatGPT also tends to oscillate between answers even when given the same question much more than humans.

Adversarial methods have also been used to present LLMs with subtle variations of a task designed to elicit mistakes for models which inadequately process semantic content. One study focusing on theory of mind tasks found that GPT-4 performed very inconsistently on false-belief tasks \citep{shapira2023clever}. On two of five task variants, GPT-4 achieved zero percent accuracy, while achieving nearly 100\% accuracy on slightly different forms of the task. The model seemed to be easily distracted by irrelevant details or make false inferences based on which information was included . Similar findings have been observed when evaluating ChatGPT on arithmetic problems, where performance degraded by 70\% when irrelevant information was included in the prompt \citep{shi2023large}.

Other studies have also found results indicating that LLMs fail to adequately represent the meaning of their prompts. An analysis of ChatGPT found that its answers to various questions about world facts or sentence meaning changed between 20\% and 40\% of the time when questions were paraphrased or reworded slightly \citep{ohmer2024form}. Similarly, an analysis of BERT models on NLI tasks showed that the models relied mostly on superficial heuristics such as presence of words like ‘all’ or ‘some’ in the premises, with performance breaking down on the subset of problems where such heuristics could not be used \citep{gubelmann2022uncovering}. These results indicate that LLMs do not truly understand the meaning of the language they process in the way a human would, raising questions about the ability to reliably generalise to unseen examples or novel domains.

\subsection{Failure to learn problem structure}

While LLMs perform well on simple instances of logical and mathematical problems, they often fail with more complex problems even when the basic steps are the same in both cases. While humans can make errors of execution due to working memory constraints or lapses of attention, LLMs fail for very different reasons. Specifically, evidence indicates that LLMs learn superficial heuristics from extensive training data while failing to learn the underlying structure of the problem or the steps needed to solve it. Memorisation of single steps observed in the training data is common, with relatively poor ability to combine multiple steps together \citep{dziri2024faith}. Various research has found that LLMs systematically fail to learn robust representations of problem structure in ways that allow them to appropriately generalise to alterations of the problem or stimulus.

LLMs are often highly sensitive to the order of answers and specific symbols used to present them. For example, one study found changes in accuracy of up to 25 percentage points depending on which order multiple choice answers were presented in \citep{alzahrani2024benchmarks}. Changing the symbols from letters to unusual characters reduced accuracies by around 5 percentage points on average. Large reductions in accuracy of 20-30 percentage points were also observed across a range of models when they were provided with irrelevant incorrect answers along with the question prompt. Another study found that the performance of Llama3 on the MMLU task fell by about 25 percentage points when model was prompted to provide its answers in a slightly different way \citep{moore2024base}.

Analysis of mathematical tasks has also found that LLMs consistently fail to learn the underlying task structure. One study used GSM-Symbolic, a maths benchmark which consisting of variants of a standard set of arithmetic questions produced by altering variable names and values \citep{mirzadeh2024gsm}. The authors found that these simple changes reduced performance by 1-9 percentage points, indicating some degree of over-fitting to the existing static dataset. They also found that including superficially important but ultimately irrelevant information reduced accuracies by up to 65 percentage points, with even o1-preview experiencing a reduction of 17 percentage points. A similar effect was observed in a separate analysis of GPT-4, where performance on a logical inference problem declined from over 95\% to about 40\% when irrelevant rules were introduced and the order of the premises was altered \citep{chen2024premise}. Such sensitivity to the alteration of variable names and inclusion of irrelevant information is not expected of a system which has adequately learned how to interpret and solve arithmetic problems.

Chain-of-thought prompting, a technique designed to help LLMs reason more carefully about complex problems, often increases LLM performance without improving generalisable reasoning capabilities. One analysis of such prompts found that for seven of the eight tasks examined, inserting a mistake near the beginning of the chain-of-thought resulted in the same answer being given more than 60\% of the time \citep{lanham2023measuring}. This indicates that much of the ‘thought’ contains post-hoc rationalisation rather than genuine reasoning. Interestingly, the authors also found this degree of post hoc reasoning was greater for larger models than smaller ones. Another analysis found that when chain-of-thought prompting was used to train the Llama model to perform an arithmetic task, performance increased from 79\% to 95\% \citep{pfau2024let}. However, when an informative task-relevant prompt was replaced by a series of meaningless filler tokens, performance only fell to 94\%, indicating that the content of the prompt was not relevant to the improved performance. Another analysis focusing on use of chain-of-thought prompting for planning found that LLMs consistently failed to generalise appropriately to new problem instances, and only performed well when given prompts highly specific and tailored to each type of problem \citep{stechly2024chain}.

One especially insightful study used a careful series of analyses to illustrate how LLMs can fail to learn the underlying structure of a task despite achieving high performance on test sets \citep{zhang2023paradox}. In a series of experiments the authors showed that the BERT model could achieve near perfect performance on a logical deduction task when tested on a subset of problems sampled in the same way as the problems on which it was trained. However, if different sampling algorithms were instead used to generate the training and testing data, performance dropped significantly, falling close to chance levels for more difficult problems. This decline in performance occurred even though both algorithms sampled from the same space of problems, with the only difference being that they did so using slightly different techniques too subtle for humans to discern any overall differences between the resulting two sets of problems. The authors also identified statistical artefacts in the training distribution, which they showed BERT had utilised to accurately perform the task when tested on data sampled in the same way as the data on which it was trained. These statistical artefacts, however, were not present when the problems were sampled in a different way, resulting in incorrect predictions when BERT asked to generalise beyond its training problems. Since such statistical artefacts are inevitable in any sample of instances of a sufficiently complex problem, it is infeasible to identify and remove all of them. This is especially significant since contemporary LLMs are so large that they can identify and utilise very complex and abstract statistical associations that may be impossible for humans to even understand. This study therefore powerfully highlights the inherent limits of the entirely statistical methods by which LLMs learn to solve abstract problems, and the difficulties in ensuring such performance will generalise to real world problems.

\subsection{The promise of reasoning models}

The past year has seen the emergence of ‘reasoning models’ such as OpenAI’s o1 and o3, DeepSeek-R1, or Google’s Gemini 2.0 Flash Thinking model. While little information about their architecture or training has been made public, these models are evidently LLMs with extensive fine-tuning on step-by-step reasoning problems taken from domains such as mathematics, coding, and science \citep{guo2025deepseek}. The models are then trained to generate a large number of candidate solutions and then score each using heuristics learned from their training data. Some have claimed that such models represent a new paradigm that will resolve the problems and limitations with traditional LLMs \citep{gold2024, todd2025teaching}. Currently, these systems are too recent for any significant evaluation of their capabilities to have been published. However, there are powerful reasons to doubt the claim that reasoning models have circumvented the problems with benchmarking and generalisation discussed in previous sections.

From a theoretical perspective, it is unclear why additional fine-tuning on step-by-step reasoning would substantially mitigate LLM limitations with generalisation or learning problem structure. As noted, reasoning models appear to work by generating large numbers of candidate chain-of-thought solutions and then using a heuristic function trained by reinforcement learning to select the most plausible candidates. Insofar as this outline is accurate, ‘reasoning models’ are not performing genuine reasoning, but are learning to mimic desired outputs by utilising heuristics and complex statistical regularities. While it may be more effective to learn to match steps in the reasoning process rather than directly predicting the final answer, the system still lacks any evident mechanism for ensuring that it learns the underlying structure of the problem rather than complex but ultimately superficial or irrelevant statistical regularities. True formal reasoning requires structural modelling of a program and performing specified operations on the component elements of the problem until the desired endpoint is reached. Based on information at the time of writing, reasoning models appear instead to be matching and mimicking reasoning steps contained in their enormous training corpus, and combining them together in accordance with complex heuristics derived from reinforcement learning. This strategy may yield performance improvements for certain types of problems, but does not resolve the key issue of the inability to learn underlying problem structure so as to facilitate robust generalisable inferences.

It has been argued that reasoning models have the potential to generate massively new training datasets that will allow for rapid improvements in performance. Specifically, the claim is that fine-tuning on step-by-step programming or mathematical reasoning tasks will enable the automatic generation of large new datasets of reasoning steps, which can then be combined with answers automatically checked for correctness \citep{todd2025teaching}. This new data can then be fed back into the model to provide even more training data, thereby further improving performance in a virtuous feedback cycle. While this approach may work in certain cases, the methodology assumes that the reasoning steps generated by model are correct and logically lead to the generated answer. This is not something that can be easily checked, and it is also inconsistent with research which indicates that chain-of-thought prompts often improve performance even while consisting largely of post-hoc reasoning that does not logically entail the generated answer \citep{lanham2023measuring, stechly2024chain}. The core problem is therefore left unsolved, namely of ensuring that the model learns the underlying logical structure of the problem and can apply this knowledge to relevant novel instances, instead of relying on superficial heuristics which will not robustly generalise to most real-world applications.

Beyond these general theoretical concerns, there are specific reasons to be cautious about the claims made about the capabilities of reasoning models. Announcements by companies like OpenAI concerning the benchmark performance of newly developed models should be interpreted as marketing designed to appeal to potential customers and investors, rather than as research reports having substantial scientific value. Too often, commentators accept such announcements entirely at face value as evidence of substantial progress, with no discussion of the extent to which such results will be reproducible, generalisable, or robust. Furthermore, in achieving these benchmarks OpenAI and other companies are likely to be making full use of any publicly available training portions of existing benchmarks. They may even be developing internal datasets with tasks resembling those from known benchmarks to use for training, thereby allowing for improved performance on those specific benchmarks which may not generalise to other tasks.

Several recent examples highlight these concerns. Recently it emerged that OpenAI was one of the major funders of the FrontierMath benchmark, and had at least some of the solutions in their possession \citep{search2025openai}. Exactly how these were used in the development of o3 is unclear, however there is more than enough reason to be suspicious about performance claims made regarding this benchmark. In addition, o3 was also trained on the public data of ARC-AI, a dataset comprised of abstract visual reasoning problems in the style of Raven’s progressive matrices \citep{chollet2024o3}. When combined with the large amount of targeted research this benchmark has attracted in recent years, the high scores achieved by o3 should not be considered a reliable metric of general reasoning capabilities. Most recently, OpenAI has claimed that o3 achieved substantially improved performance on the CodeForces benchmark and International Olympiad in Informatics 2024 competition \citep{openai2025competitive}. While the questions from these evaluations had already been made public by the time they were used for assessing o3, OpenAI claims their training cutoff was after the release of these problems, and that manual search revealed no evidence of contamination of the training data. They also claimed that the performance increase was due entirely to reinforcement learning of step-by-step reasoning, with no specific tailoring to the specific benchmarks. Though at present these claims are impossible to verify, even if taken at face value, information from the benchmarks could still have influenced the development of o3 through selection of hyperparameters or other development decisions. As an illustration of how this can occur, one study using older LLMs illustrated how information from a test dataset can be `illicitly' learned by a series of seemingly benign intermediate evaluation steps \citep{mansurov2024data}. As such, until OpenAI's claims can be independently verified using novel problems, it should not be assumed that the claimed improvements in benchmark scores will translate into an equivalently large and robust improvement in generalised capabilities.

In the case of the recently-released DeepSeek-R1, although more information is available than for OpenAI's o3 model, it is still highly unclear what data was used for training \citep{guo2025deepseek}. The authors mention a combination of synthetic and human-annotated data used for both chain-of-thought fine-tuning and reinforcement learning, but no details are provided as to the nature of the data or what types of tasks were trained on. It is rumoured that it was trained on outputs generated by OpenAI's o1 model, which calls into question how generalisable its reasoning capabilities really are when applied to novel types of problems not reflected in its training data. DeepSeek-R1 has been made publicly available, which should facilitate community analysis of the robustness of its general reasoning cognitive capabilities.

Overall, there are strong theoretical and empirical reasons to doubt that AI reasoning models have substantially resolved the limitations of previous generations of LLMs. New models like o3 still rely on the same underlying mechanisms that extract superficial heuristics from huge datasets without learning the structure of a problem. Likewise, state-of-the-art scores on benchmarks which were already known and likely guided o3’s development are not a reliable indicator of its capabilities of genuine reasoning or generalisation to novel tasks. 

\section{Conclusion}

It is undeniable that recent years have seen substantial progress in the development of large language models capable of performing many tasks relating to language, logic, coding, and mathematics. However, the extent of this progress is frequently exaggerated based on appeals to rapid increases in performance on various benchmarks. I have argued that these benchmarks are of limited value for measuring LLM progress because of problems of models being over-fit to the benchmarks, lack real-world relevance of test items, and inadequate validation for whether the benchmarks predict general cognitive performance. Conversely, evidence from adversarial tasks and interpretability research indicates that LLMs consistently fail to learn the underlying structure of the tasks they are trained on, instead relying on complex statistical associations and heuristics which enable good performance on test benchmarks but generalise poorly to many real-world tasks. The result is that the LLMs lack robust reasoning capabilities that are consistent and invariant to irrelevant changes in problem format, style, numerical data, or context. Despite much recent hype, new reasoning models do not offer a general solution to these problems. Renewed hype highlights the importance of subjecting new LLMs to careful systematic evaluation to determine its reasoning capabilities and their limits across a range of scenarios. Benchmark performance alone is insufficient to establish general reasoning competence.

\bibliographystyle{unsrt}
\bibliography{references}  

\end{document}